\documentclass[a4]{article}
\usepackage[fleqn]{amsmath}
\usepackage{graphicx}
\usepackage{bm}
\title{Analysis of Dropout in Online Learning}
 \author{Kazuyuki Hara}
 \date{\today}
\begin{document}
\maketitle
\begin{description}
\item[abstract]
Deep learning is the state-of-the-art in fields such as visual object recognition and speech recognition. This learning uses a large number of layers and a huge number of units and connections. Therefore, overfitting is a serious problem with it, and the dropout which is a kind of regularization tool is used. However, in online learning,  the effect of dropout is not well known. This paper presents our investigation on the effect of dropout in online learning. We analyzed the effect of dropout on convergence speed near the singular point. Our results indicated that dropout is effective in online learning. Dropout tends to avoid the singular point for convergence speed near that point. 
\end{description}

\section{Introduction}
Deep learning \cite{Hinton2006,LeCun2015} is attracting much attention in the fields of visual object recognition, speech recognition, object detection, among many others. It provides automatic feature extraction, and it can achieve outstanding performance \cite{Hinton2012a,Deng2013}. 

Deep learning uses a deep layered network and a huge number of data, so overfitting is a serious problem with it. Regularization is used to avoid overfitting. Hinton et al. proposed a regularization method called ``dropout" \cite{Hinton2012} for this purpose. Dropout follows two processes. During learning, some hidden units are randomly removed from a pool of hidden units with a probability $q$, thereby reducing the network size. During evaluation, the output of the learned hidden units and the output of those not learned are summed up and multiplied by $p=1-q$. Hinton pointed out that dropout has some effect on ensemble learning. Baldi et al. theoretically analyzed dropout as ensemble learning\cite{Baldi2013}. Warger et al. also theoretically analyzed dropout as an adaptive L2 regularizer\cite{Wager2014}. However, their analysis method is very different. 

This paper presents our analysis of the dropout in online learning, which is not well studied. Online learning may be useful in deep networks, where a huge amount of data and a very large number of network parameters are required. However, how the dropout is effective in online learning is not known. In this paper, we utilized a multilayer perceptron because a simple network is suitable for precisely investigating the effect of dropout\cite{Biehl1995,Saad1995}. We investigated two points: the behavior of the network using dropout and the behavior near the singular point \cite{hara2017}. We determined the effect of dropout in on-line learning using computer simulations.

\section{Model}
\label{model}
\subsection{Network Structure}
In this paper, we use a teacher-student formulation and assume the existence of a teacher network (teacher) that produces the desired output for the student network (student). By introducing the teacher, we can directly measure the similarity of the student input-to-hidden weight vector to that of the teacher. First, we formulate a teacher and a student and then introduce the gradient descent algorithm. 

\begin{figure}[h]
\begin{center}
\includegraphics[width=7cm]{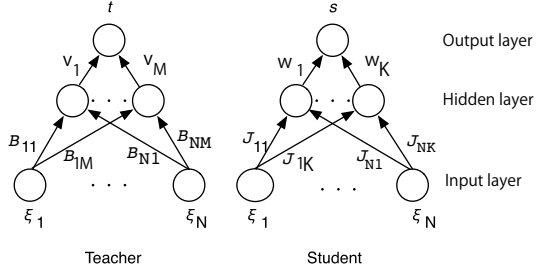}
\end{center}
\caption{Network structures of teacher and student}
\label{network}
\end{figure}

The teacher and student are a three-layer perceptron (MLP) with $N$ input units, some hidden units, and an output, as shown in Fig. \ref{network}. The teacher consists of $M$ hidden units, and the student consists of $K$ hidden units. Each hidden unit is a perceptron. The $n$th input-to-hidden weight vector of the teacher is $\bm{B}_n=(B_{n1}, \ldots, B_{nN})$, and the $i$th input-to-hidden weight vector of the student is $\bm{J}_{i}^{(m)} = (J_{i1}^{(m)}, \ldots, J_{iN}^{(m)})$, where $m$ denotes the learning iterations. In the MLP, all hidden-to-output weights for teacher $\bm{v}$ are fixed some values, and those for student $\bm{w}$ are learnable\cite{Park2005}. 

We assume that the $k$th elements $\xi_k^{(m)}$ of the independently drawn input $\bm{\xi}^{(m)}=(\xi_1^{(m)}, \ldots, \xi_N^{(m)})$ are uncorrelated random variables with zero mean and unit variance; that is, the $k$th element of the input is drawn from a probability distribution $\mbox{P}(\xi_k)$. The thermodynamic limit of $N \rightarrow \infty$ is also assumed. The statistics of the inputs in the thermodynamic limit are $\left<\xi_{k}^{(m)}\right> =0$, $\left<(\xi_{k}^{(m)})^2\right>\equiv \sigma_{\xi}^2=1$, and $\left<\|\bm{\xi}^{(m)}\|\right>=\sqrt{N}$, where $\left< \cdots \right>$ denotes the average and where $\| \cdot \|$ denotes the norm of a vector. 

We assume that both the teacher and the student receive $N$-dimensional input $\bm{\xi}^{(m)}$, so the teacher output $t^{(m)}$ and the student output $s^{(m)}$ are

\begin{align}
t^{(m)}=&\sum_{n=1}^{M}v_n t_n^{(m)}=\sum_{n=1}^{M}v_n g(d_n^{(m)}),\\
s^{(m)}=&\sum_{i=1}^{K} w_{i} s_{i}^{(m)}=\sum_{i=1}^{K} w_{i} g(y_{i}^{(m)}).
\end{align}

Here, $v_n$ and $w_{i}$ are the hidden-to-output weight of the teacher and the student, respectively. $g(\cdot)$ is the activate function of a hidden unit, and $d_n^{(m)}$ is the inner potential of the $n$th hidden unit of the teacher calculated using 

\begin{equation}
d_n^{(m)}=\sum_{k=1}^N B_{nk} \xi_k^{(m)}.
\end{equation}

\noindent
$y_{i}^{(m)}$ is the inner potential of the $k'$th hidden unit of the student calculated using 

\begin{equation}
y_{i}^{(m)}=\sum_{k=1}^N J_{ik}^{(m)} \xi_{k}^{(m)}. 
\end{equation}

Next, we show the teacher network weight settings. The hidden-to-output weight $v_n$ is set to 0.5. Each element of input-to-hidden weight $B_{nk}, \ n=1 \sim K$ is drawn from a probability distribution with zero mean and $1/N$ variance. With the assumption of the thermodynamic limit, the statistics of the teacher input-to-hidden weight vector are $\left<B_{nk}\right>=0, \left< (B_{nk})^2\right> \equiv \sigma_{B}^2=1/N$, and $\left<\| \bm{B_n}\|\right>=1$. This means that any combination of $\bm{B}_{n} \cdot \bm{B}_{m}=0$. The distribution of inner potential $d_{n}^{(m)}$ follows a Gaussian distribution with zero mean and unit variance in the thermodynamic limit. 

After that, we show the student network weight settings. The initial value of the hidden-to-output weight $w_{i} ^{(0)}$ is set to zero mean and to 0.1 variance. For the sake of analysis, we assume that each element of the input-to-hidden weight $J_{ik}^{(0)}$, which is the initial value of the student vector $\bm{J}_{i}^{(0)}$, is drawn from a probability distribution with zero mean and with $1/N$ variance. The statistics of the $i$th input-to-hidden weight vector of the student are $\left< J_{ik}^{(0)}\right>=0, \left<(J_{ik}^{(0)})^2\right>\equiv \sigma_{J}^2=1/N$, and $\left<\|\bm{J}_{i}^{(0)}\|\right>=1$ in the thermodynamic limit. This means that any combination of $\bm{J}_{i}^{(0)} \cdot \bm{J}_{j}^{(0)}=0$. The activate function of the hidden units of the student $g(\cdot)$ is the same as that of the teacher. The statistics of the student input-to-hidden weight vector at the $m$th iteration are $\left< J_{ik}^{(m)}\right>=0$, $\left<(J_{ik}^{(m)})^2\right> =Q_{ii}^{(m)}/N$, and $\left<\|\bm{J}_{i}^{(m)}\|\right>=\sqrt{Q_{ii}^{(m)}}$. Here, $Q_{ij}^{(m)}=\bm{J}_{i}^{(m)}\cdot \bm{J}_{j}^{(m)}$. The distribution of the inner potential $y_{i}^{(m)}$ follows a Gaussian distribution with zero mean and $Q_{ij}^{(m)}$ variance in the thermodynamic limit. 

\subsection{Learning algorithm}
Next, we introduce the stochastic gradient descent (SGD) algorithm for the soft committee machine. The generalization error ($\epsilon_g$) is defined as the squared error $\varepsilon$ averaged over possible inputs that are independent of learning data\cite{Park2005}: 

\begin{align}
\epsilon_g^{(m)} &=\left<\epsilon^{(m)} \right>=\frac{1}{2}\left< (t^{(m)}-s^{(m)})^2 \right>\nonumber \\
&=\frac{1}{2}\left< \left(\sum_{n=1}^K v_n g(d_n^{(m)})-\sum_{i=1}^{K} w_{i}^{(m)} g(y_{i}^{(m)})\right)^2 \right> \\
&=\frac{1}{\pi}\left[\sum_{n=1}^Mv_n^2\arcsin\left(\frac{1}{2}\right)+\sum_{i=1}^K (w_i^{(m)})^2\arcsin\left(\frac{Q_{ii}^{(m)}}{1+Q_{ii}^{(m)}}\right)\right.\nonumber \\
&+2\sum_{i=1}^K\sum_{j>i}^K w_j^{(m)}w_j^{(m)}\arcsin\left(\frac{Q_{ij}^{(m)}}{\sqrt{1+Q_{ii}^{(m)}}\sqrt{1+Q_{jj}^{(m)}}}\right) -2\left\{\sum_{i=1}^Mw_i^{(m)}v_i\arcsin\left(\frac{R_{ii}^{(m)}}{\sqrt{2(1+Q_{ii}^{(m))}}}\right)\right.\nonumber\\
&+ \left.\left.\sum_{n=1}^{M}\sum_{i\neq n}^K w_i^{(m)}v_n\arcsin\left(\frac{R_{in}^{(m)}}{\sqrt{2(1+Q_{ii}^{(m)}})}\right)\right\}\right].
\label{eg}
\end{align}

\noindent
Here, we assume that $T_{nn}=1$ and $T_{nm}=0$, where $n\neq m$. This equation shows that if $M$ of $Q_{ii}$s and $R_{ii}$s converge into 1, the rest of $Q_{ii}$s, $R_{ii}$s, $Q_{ij}$s, and $R_{in}$s vanish, and $M$ of $w_{i}$ is equal to $v_n$ at $t\rightarrow \infty$; then, the generalization error becomes zero. Here, we denote that $Q_{ii}=\bm{J}_i \cdot \bm{J}_i$, $R_{ii}=\bm{B}_i \cdot \bm{J}_i$, $Q_{ij}=\bm{J}_i\cdot \bm{J}_j$, and $R_{in}=\bm{B}_n \cdot \bm{J}_i$. 

As already described, when the teacher consists of $M$ hidden units and the student consists of $K$ hidden units, three cases emerge for setting the number of hidden units in the students: (1) $M> K$, (2) $M= K$, and (3) $M<K$. The case of $M>K$ is unlearnable and insufficient because the degree of complexity of the students is less than that of the teacher. The case of $M=K$ is learnable because the degree of complexity of the students is the same as that of the teacher. The case of $M<K$ is learnable and redundant because the degree of complexity of the students is higher than that of the teacher \cite{Bishop}.

Next, we show the learning equations\cite{Biehl1995,Park2005}. At each learning step $m$, a new uncorrelated input, $\bm{\xi}^{(m)}$, is presented, and the current hidden-to-output weight $w_{i}^{(m)}$ and that of the input-to-hidden weight of the student $\bm{J}_{i}^{(m)}$ are updated using

\begin{align}
\bm{J}_{i}^{(m+1)}&=\bm{J}_{i}^{(m)}+\frac{\eta}{N}\delta_i^{(m)} w_{i}g^{\prime}(y_{i}^{(m)})\bm{\xi}^{(m)}, \label{le}\\
\bm{w}_{i}^{(m+1)}&=\bm{w}_{i}^{(m)}+\frac{\eta}{N}\delta_i^{(m)} y_{i}, \\
\delta_i^{(m)}&=\sum_{n=1}^M v_n g(d_n^{(m)})-\sum_{j=1}^{K}w_{j}g(y_{j}^{(m)}),
\end{align}

\noindent 
where $\eta$ is the learning step size and where $g^{\prime}(x)$ is the derivative of the activate function of the hidden unit $g(x)$. 

\begin{figure}[ht]
\begin{center}
\includegraphics[width=6cm,height=2.5cm]{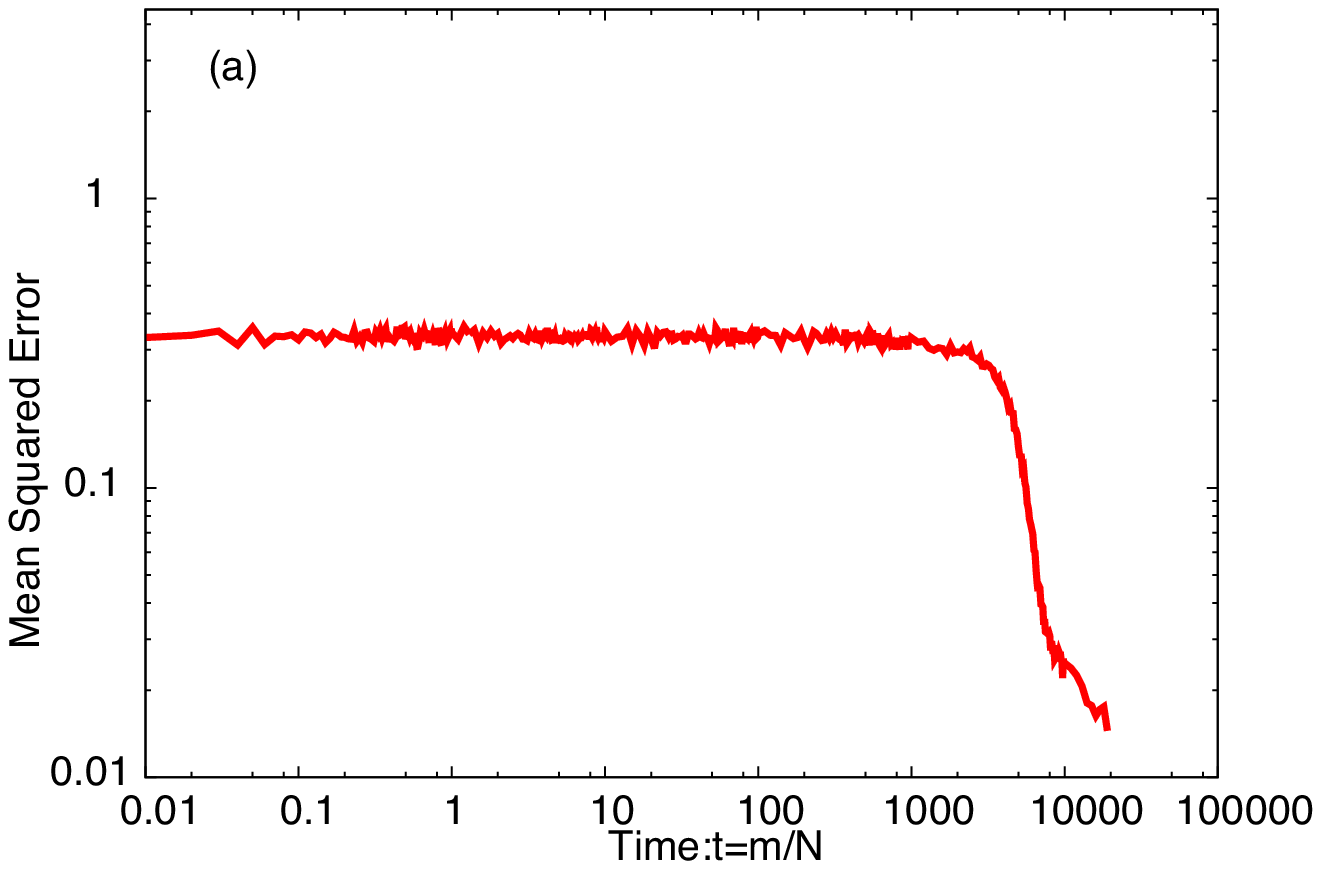}\\
\includegraphics[width=5.6cm,height=2.4cm]{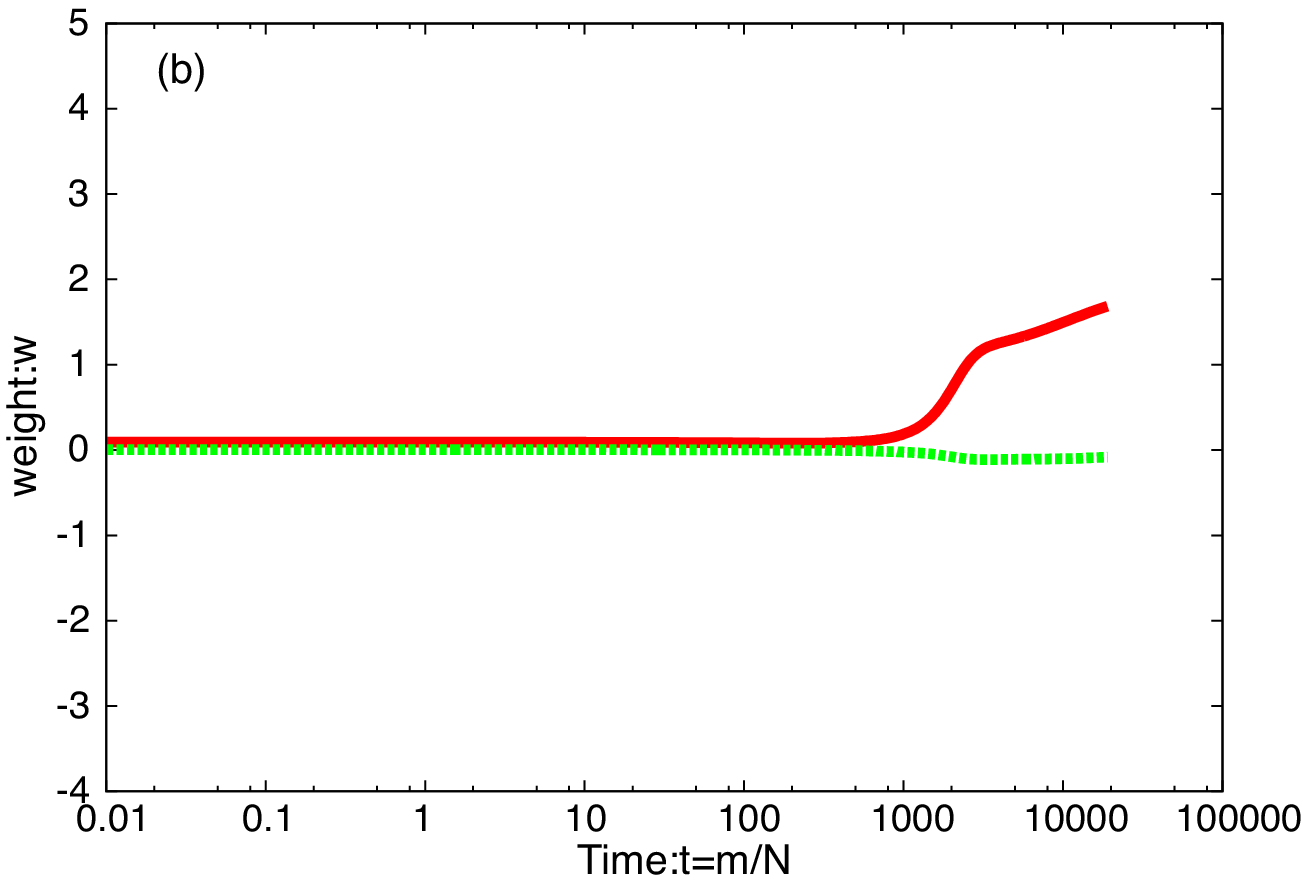}\\\includegraphics[width=6cm,height=4cm]{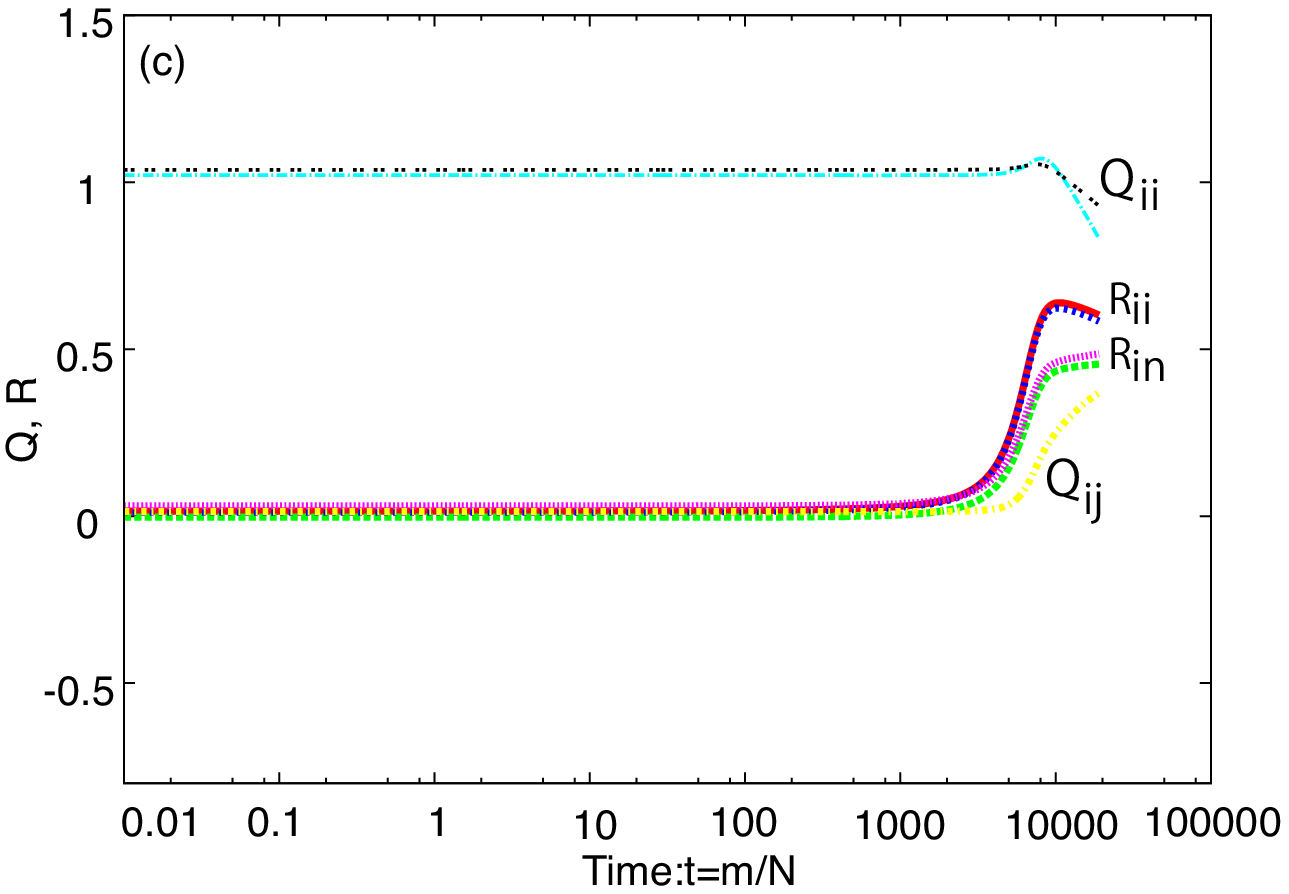}
\end{center}
\caption{\label{mse}Dynamic behavior of SGD with $M=K=2$. (a) shows MSE, (b) shows $\bm{w}$, and (c) shows $Q$ and $R$.}
\end{figure}

\begin{figure}[ht]
\begin{center}
\includegraphics[width=6cm,height=2.5cm]{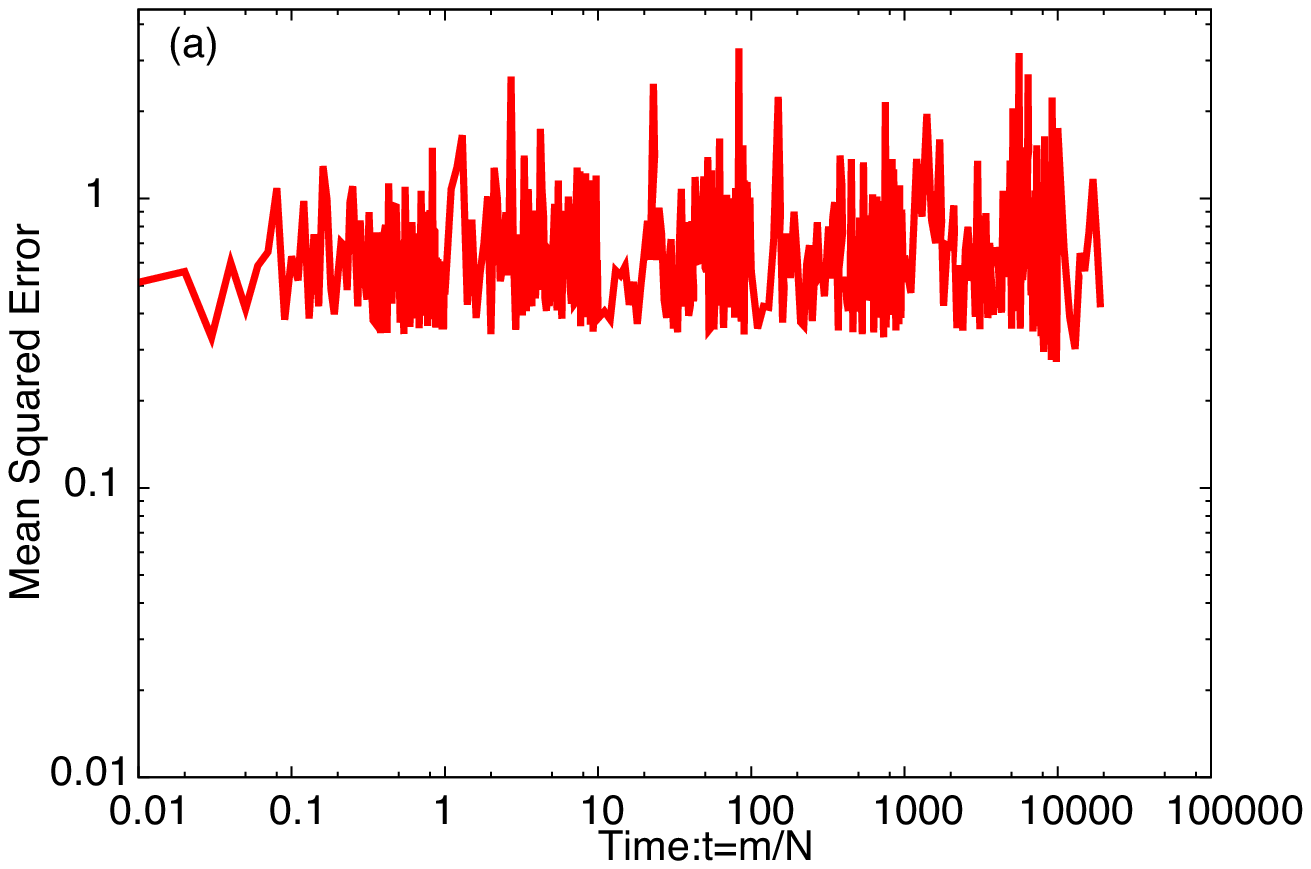}\\
\includegraphics[width=5.6cm,height=2.4cm]{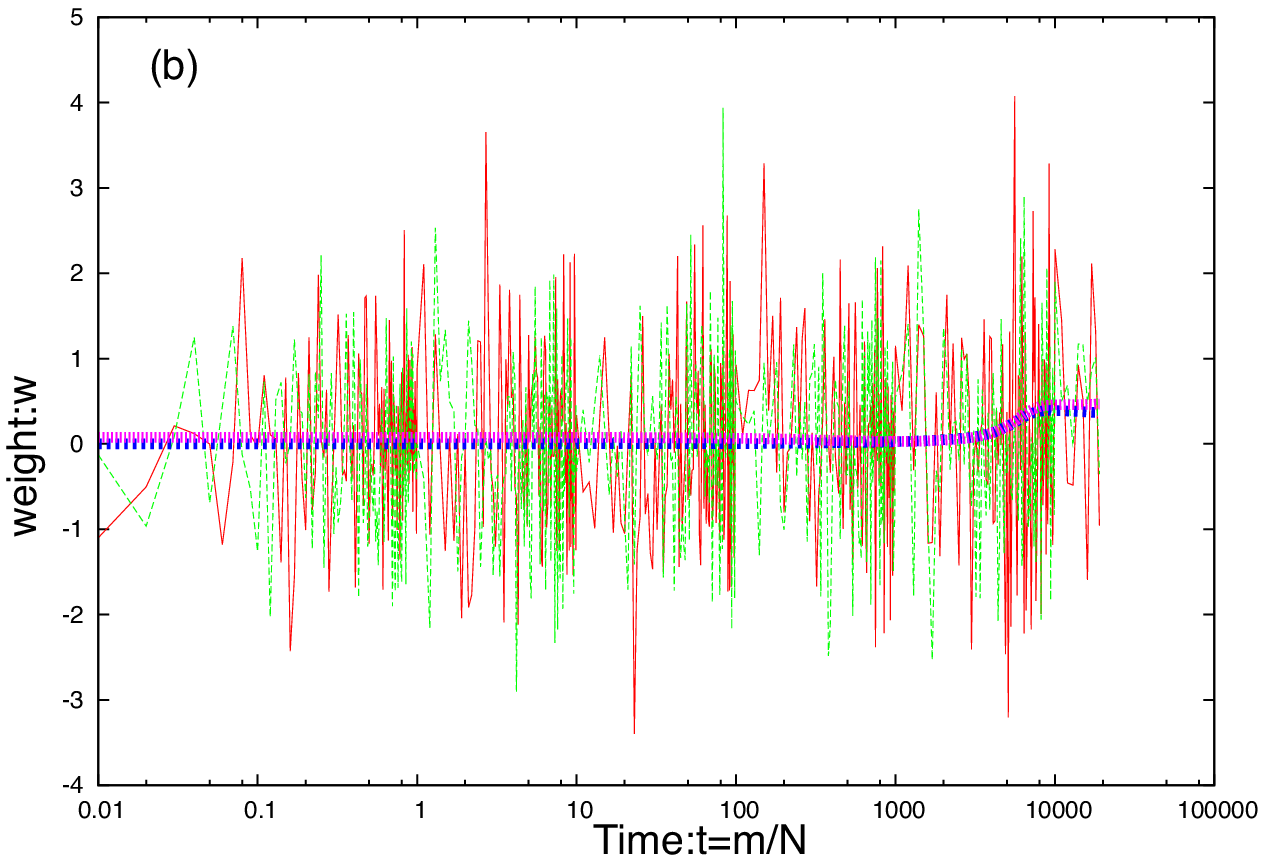}\\\includegraphics[width=6cm,height=4cm]{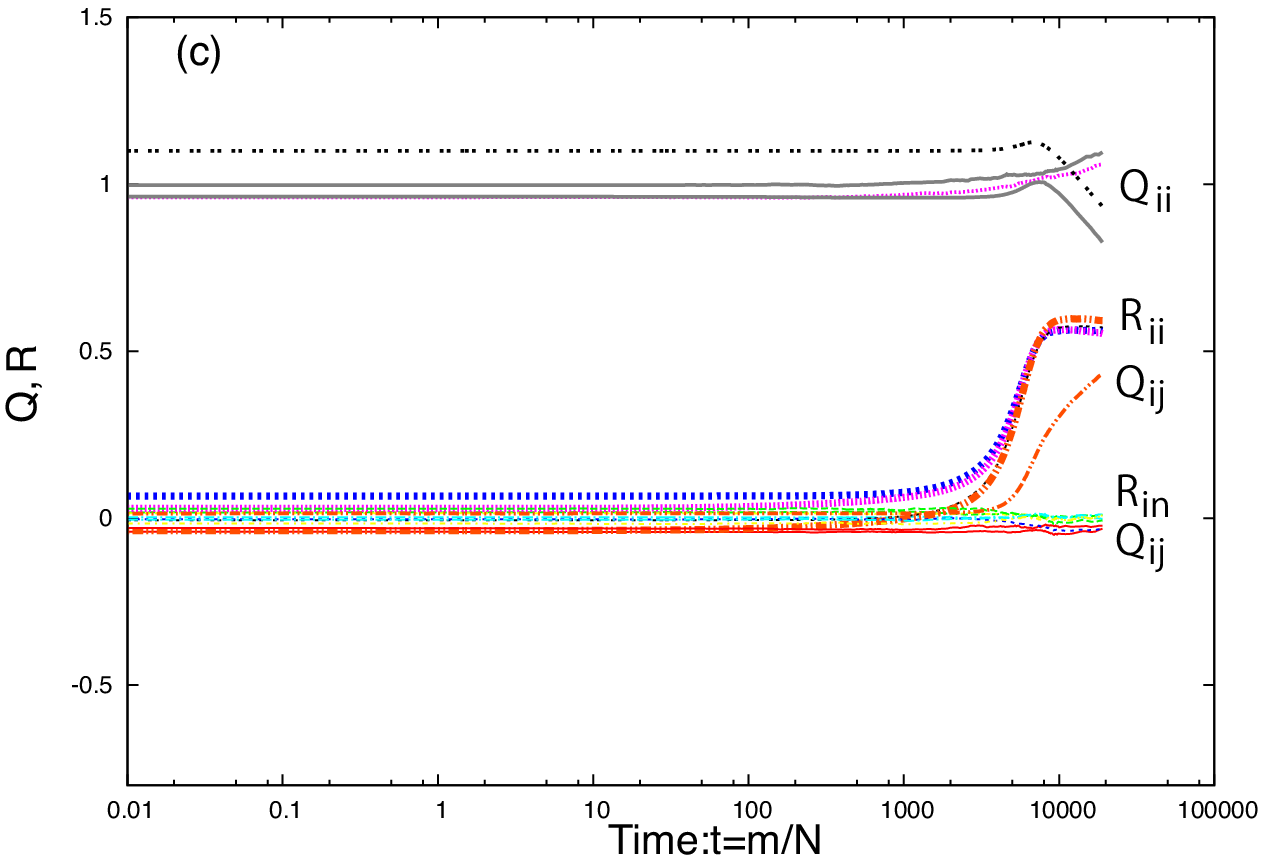}
\end{center}
\caption{\label{mse2}Dynamic behavior of SGD with $M=2$ and $K=4$. Top shows MSE, middle shows $\bm{w}$, and bottom shows $Q$ and $R$.}
\end{figure}

Next, we show the typical behavior of learning using SGD in computer simulations. The teacher is set as described in the previous subsection. The student is initialized as described in the previous subsection. We use two settings of (1)$M=2$ and $K=2$ (Fig. \ref{mse}), and (2) $M=2$ and $K=4$ (Fig. \ref{mse2}). (1) is a learnable setting and (2) is a redundant setting. The activate function $g(x)$ is the sigmoid function, $\mbox{erf}(x/\sqrt{2})$. The number of input units $N$ is set to $1000$, and the learning step size is set to $\eta=0.005$. In these figures, the horizontal axis is time $t=m/N$. In Fig. \ref{mse}(a), we show the time course of the mean squared error (MSE), Fig.\ref{mse}(b) show that of $\bm{w}$, and Fig.\ref{mse}(c) show that of $Q_{ii}$ and $Q_{ij}$ (referred to as $Q$s) and $R_{ii}$ and $R_{in}$ (referred to as $R$s).


\noindent
Figure \ref{mse} shows the results for learnable case (case (1)). Figure \ref{mse}(a) shows two states in the learning process: one is a plateau, and the other is symmetry breaking of the weights. The plateau is a phenomenon where the MSE decreases very slowly for a long interval in the learning process. Symmetry breaking of the weights is a phenomenon involving a sudden decrease in MSE because of the success of credit assignment of the hidden units. To achieve $\epsilon_g \rightarrow 0$ in case (1), as we showed in Eq. \ref{eg}, $R_{ii}s \rightarrow 1$, $Q_{ii}s \rightarrow 1$, $R_{in}s \rightarrow 0$, and $Q_{ij}s \rightarrow 0$ at the limit of $t\rightarrow \infty$. From Fig. \ref{mse}, MSE decreases properly, and $\bm{w}$ converges into $\bm{w}=(2,0)$ while $\bm{v}=(0.5,0.5)$. For $R$s, $R_{ii}s > R_{in}s$ and is more than 0.5, and $R_{in}$ does not vanish. $Q_{ii}s \sim 1$, and $Q_{ij}s \rightarrow 0.5$; thus, $Q_{ij}s$ also does not vanish. These results show that the teacher and student converge at different parameters. Note that in this case, the dynamics of $\bm{w}$ are stable. 

Figure \ref{mse2} shows the results for case (2). As aforementioned, the teacher is $M=2$, and the students are $K=4$. Then, two elements of $\bm{w}$ and two of $\bm{J}_i$ are redundant. To achieve $\epsilon_g \rightarrow 0$ in this case, the same conditions are required as those in case (1), and in addition, $K-M$ of the elements of $\bm{w}$ and $K-M$ of $Q_{ii}$s must vanish. This means that $K-M$ of $\bm{J}_i$ and $K-M$ elements of $\bm{w}$ must vanish. From Fig. \ref{mse2}(a), MSE becomes bumpy, and it does not decrease properly. We cannot see clearly plateau and symmetry breaking of the weights. Figure \ref{mse2}(b) shows the time course of $\bm{w}$. In the figure, two thick lines in the middle are shown to converge properly into (0,5,0.5), while the rest are bumped (shown by thin lines). This phenomenon occurs because two of $\bm{w}$ are redundant, and no constraint is used to eliminate redundant weights. Figure \ref{mse2}(c) shows the time course of $Q$s and $R$s. This figure shows that $R_{in}s$ and $Q_{ij}s$ almost vanish. Two of $Q_{ii}s$ stay at $Q_{ii}=1$, and the rest seems to decrease properly to $Q_{ii}<1$. However, two of $R_{ii}$ and one of $R_{in}s$ converge into 0.6, while two of $R_{ii}s$ should converge into 1.0, and $R_{ins}$ must vanish. Also, one of $Q_{ij}s$ becomes a larger value at $t=20,000$; however, it should vanish. These will cause a large MSE. Note that the MSE does not decrease properly, but Rs and Qs are updated properly. This fact shows that the MSE is not a good enough index of the learning performance and that the teacher-student formulation gives additional information. 

\section{Dropout}
\label{dropout}
In this section, we introduce dropout \cite{Hinton2012} and its behavior in online learning\cite{hara2017}. Dropout is used in deep learning to prevent overfitting. A small number of data compared with the number of units and weights of a network may cause overfitting \cite{Bishop}. In the state of overfitting, the learning error (the error for learning data) and the test error (the error given by cross-validation) become different. This means that the error for learning data is small; however, the error for overall data is large. In this paper, we assume dropout is carried out in online learning. 

The learning equation of dropout or a MLP can be written as follows\cite{hara2017}.

\begin{align}
\bm{J}_{i}^{(m+1)}=&\bm{J}_{k'}^{(m)}+\frac{\eta}{N}\delta_i^{D} w_i^{(m)} g^{\prime}(y_{i}^{(m)}) \bm{\xi}^{(m)}, \label{led}\\
\delta_i^{D}=& \sum_{n=1}^M v_n g( d_n^{(m)})-\sum_{j\in D^{(m)}}^{pK}w_{j} g(y_{j}^{(m)}).
\label{delta_D}
\end{align}

\noindent
Here, $D^{(m)}$ shows a set of the hidden units that are randomly selected with respect to the probability $p$ from all the hidden units at the $m$th iteration. Note that the second term of the right hand side in Eq. (\ref{delta_D}) is the MLP output composed of selected hidden units. The hidden units in $D^{(m)}$ are subject to learning, and the size of the student decreases due to dropout. After the learning, the student's output $s^{(m)}$ is calculated using the sum of learned hidden outputs and hidden outputs that have not been learned multiplied by $p$\cite{hara2017}.

\begin{equation}
s^{(m)}=p*\left\{\sum_{i\in {D}^{(m)}}^{pK} w_{i} g(y_{i}^{(m)})+\sum_{j \notin D^{(m)}}^{(1-p)K} w_{j} g(y_{j}^{(m-1)})\right\}
\label{dropout_learning}
\end{equation}

\noindent
This equation is regarded as the ensemble of a learned network (written by $y_i^{(m)}$) and that of a not learned network (written by $y_{j}^{(m-1)}$) when the probability is $p = 0.5$. However, in deep learning, selected hidden units in $D^{(m)}$ are changed at every iteration, where the same set of hidden units are used in the ensemble learning. Therefore, dropout is regarded as ensemble learning using a different set of hidden units at every iteration\cite{hara2017}. 

\subsection{Analytical Results}
\begin{figure}[ht]
\begin{center}
\includegraphics[width=6cm,height=2.5cm]{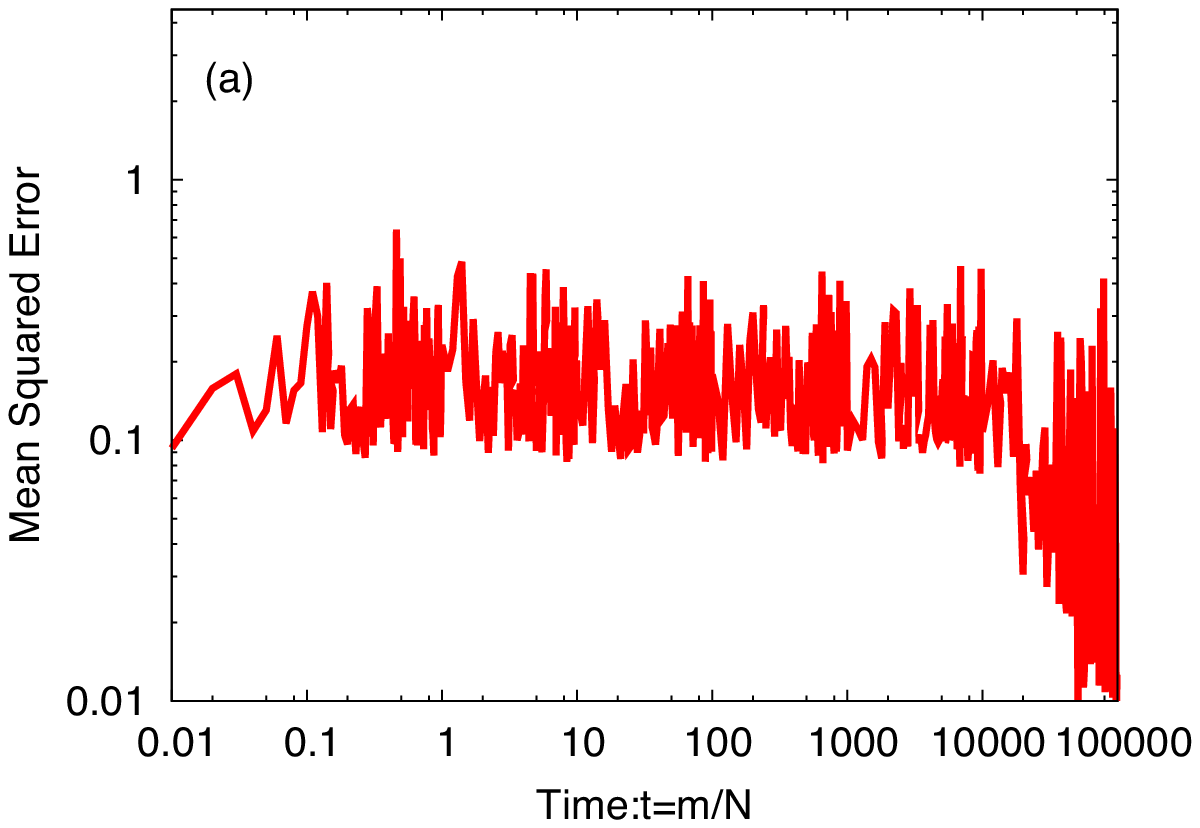}\\
\includegraphics[width=5.3cm,height=2.4cm]{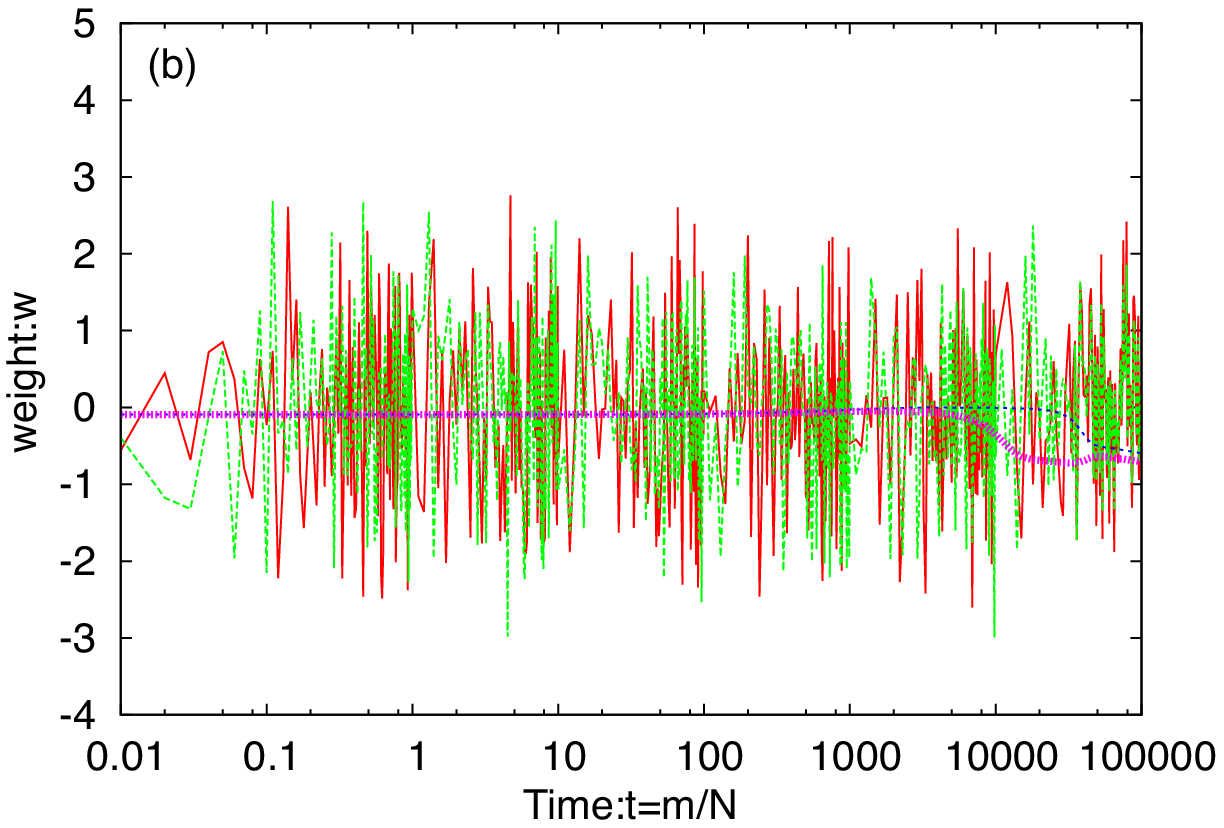}\\
\includegraphics[width=6cm,height=4cm]{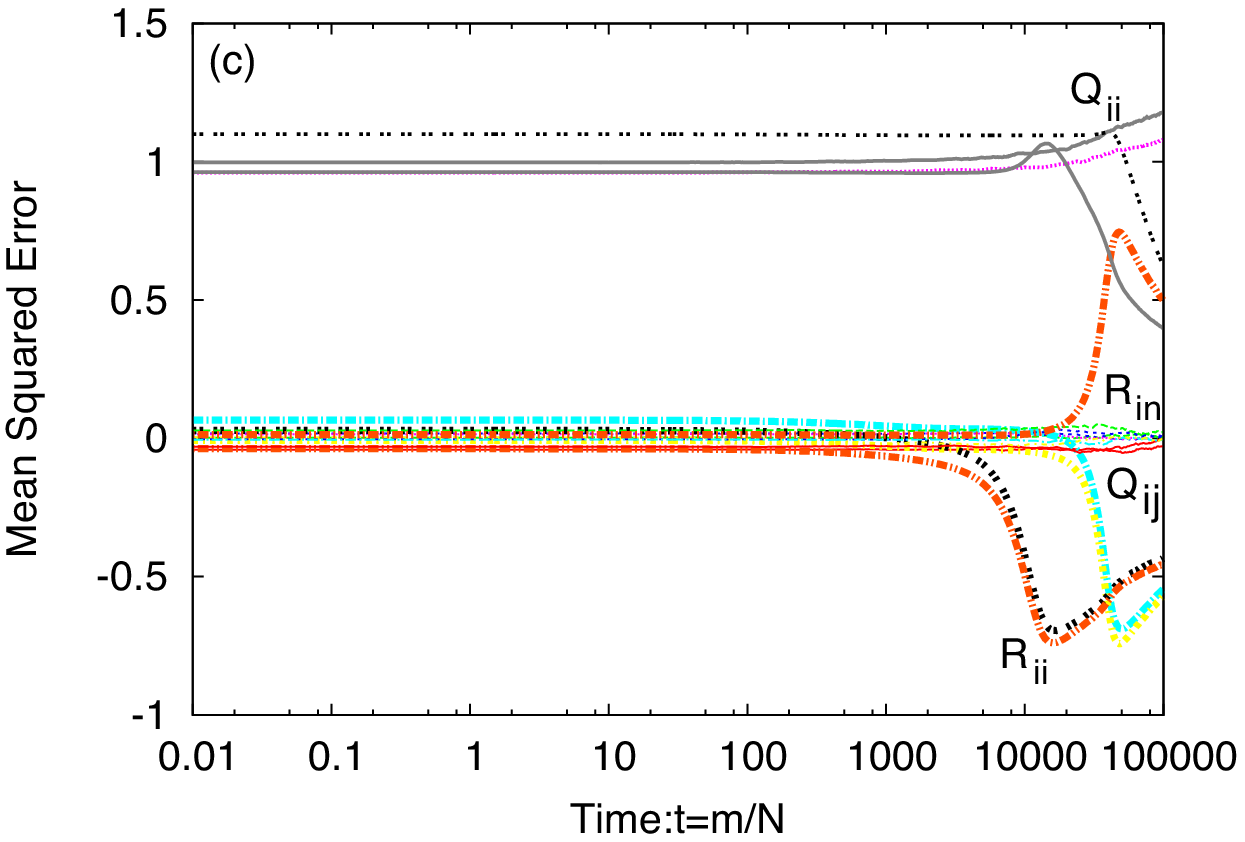}
\end{center}
\caption{\label{mse3}Dynamic behavior of dropout with $M=2$ and $K=4$. (a) shows MSE, (b) shows $\bm{w}$, and (c) shows $Qs$ and $Rs$.}
\end{figure}

We show the results of analysis of the dropout in on-line learning through computer simulations. Figure \ref{mse3} shows the time course of the MSE, that of the hidden-output weight $\bm{w}$, and that of $R$s and $Q$s. Dropout with SGD is referred to as dropout in this paper. The computer simulation conditions are as the same as those in Fig. \ref{mse}. In figure \ref{mse3}, the horizontal axis is time $t=m/N$. 

In Fig. \ref{mse3}(a), the MSE is as bumpy as the one in Fig. \ref{mse2}(a), but the baseline of MSE decreased to 0.1. Symmetry breaking of the weights is also observed when $t>20,000$. From Fig. \ref{mse3}(b), the time course of two of $\bm{w}$, shown by thick lines, properly converged to (-1,-1) . The rest of $\bm{w}$, shown by thin lines, behaves as bumpy as shown in the figure. The student has four $Q_{ii}$s and six $Q_{ij}$s. Fig. \ref{mse3}(c) shows that two of $Q_{ii}s$ stayed at $Q_{ii}=1$, and the other two decreased to about 0.5 at $t=100,000$. This mean that two weights in the student have the same norm as that of the teacher, and the remaining two weights will vanish. This phenomenon may cause symmetry breaking of the weights. These results show that dropout tends to decrease the MSE when the student is redundant by eliminating the redundant weights. 

\subsection{Singular Teacher}
H. Park pointed out that for a singular case, slow dynamics are observed\cite{Park2005}. Thus, we investigated the effect of dropout when the teacher is singular. We set the teacher ($M=2$) as follows. The input-to-output weights were set to $\bm{B}_1=\bm{B}_2$, and the hidden-to-output weights were set to $v_1=v_2=0.5$. H. Park also pointed out that a quasi plateau is caused by the singular subspace ($w_1+w_2=1$ for $K=2$), which does not exist in the soft-committee machine\cite{Park2005}. Thus, $v_1=v_2=0.5$ will induce the student to fall into this singular subspace. 

\begin{figure}[ht]
\begin{center}
\includegraphics[width=6cm,height=2.5cm]{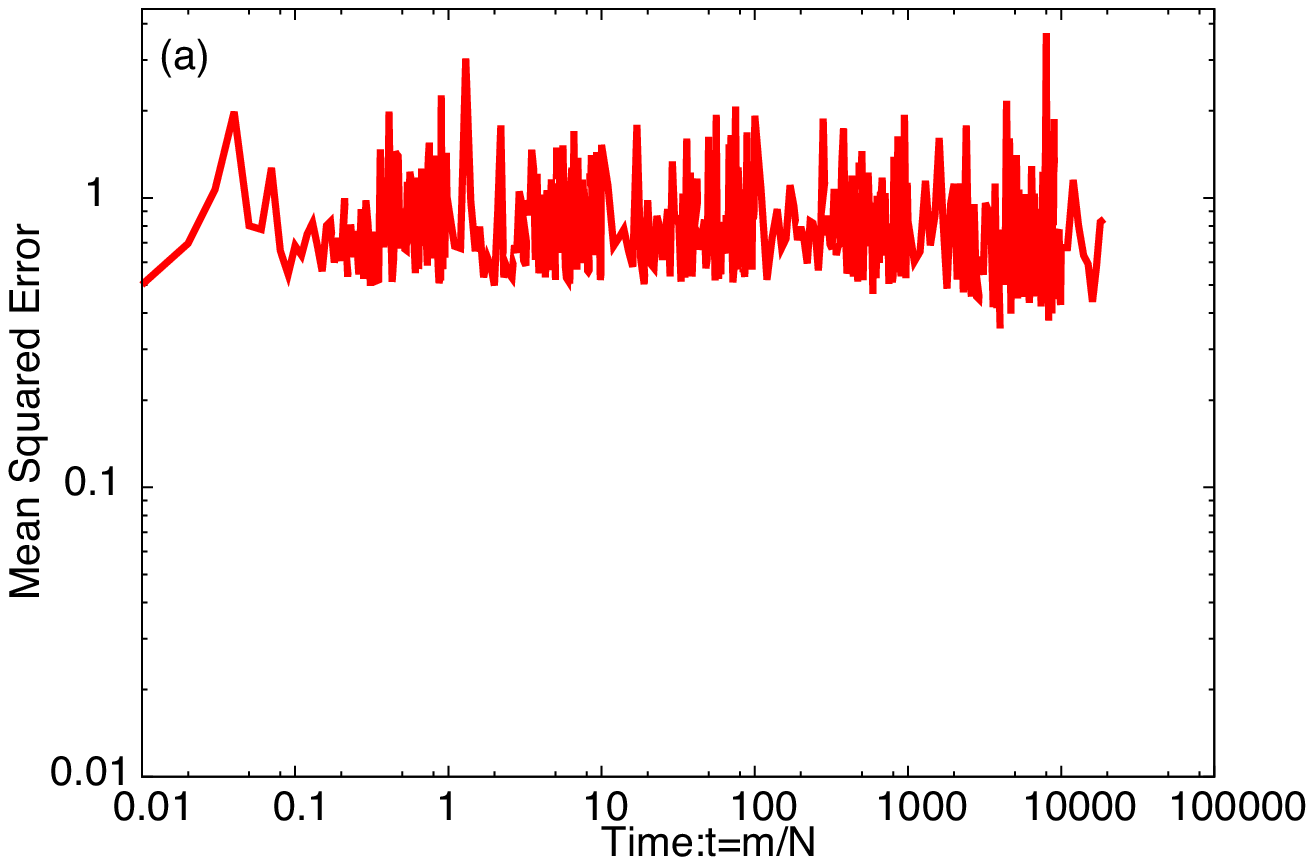}\\
\hspace{0.1cm}\includegraphics[width=6cm,height=2.4cm]{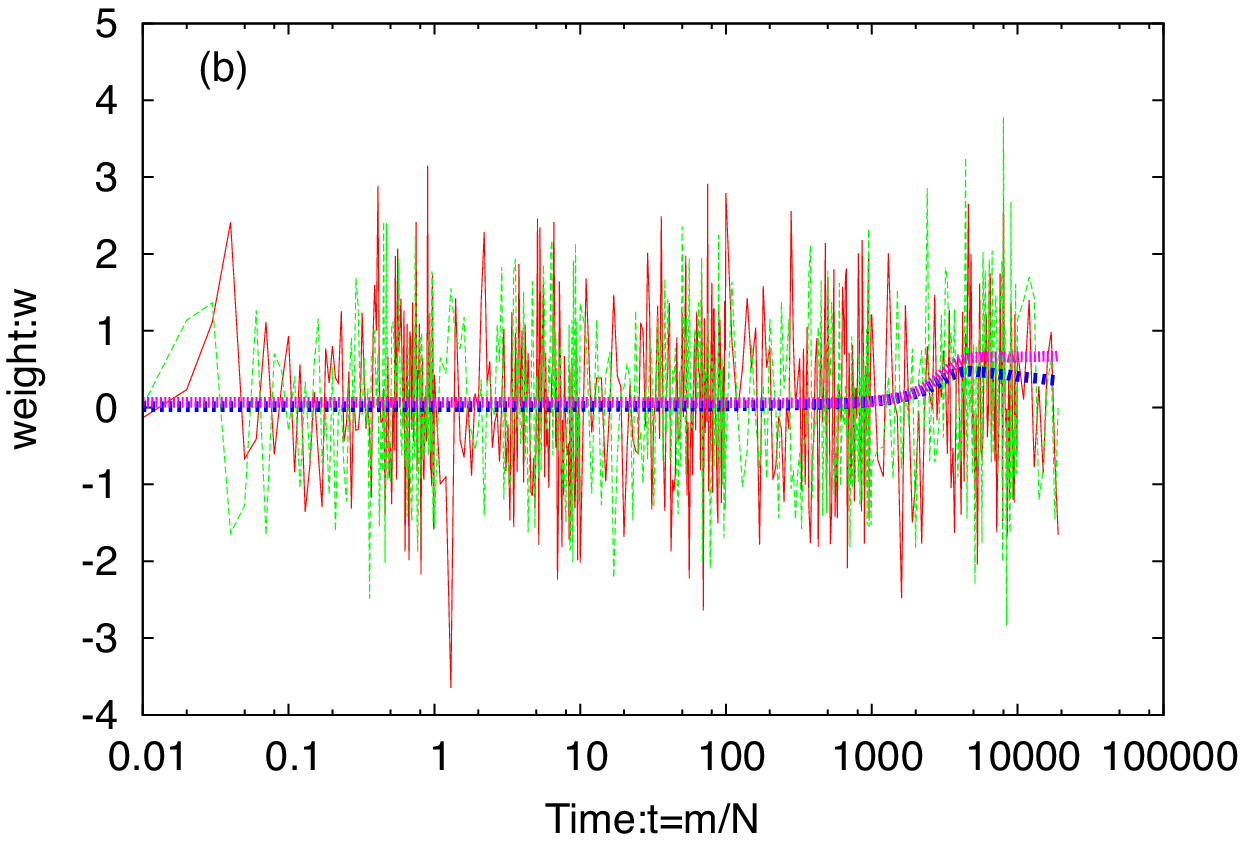}\\\includegraphics[width=6cm,height=4cm]{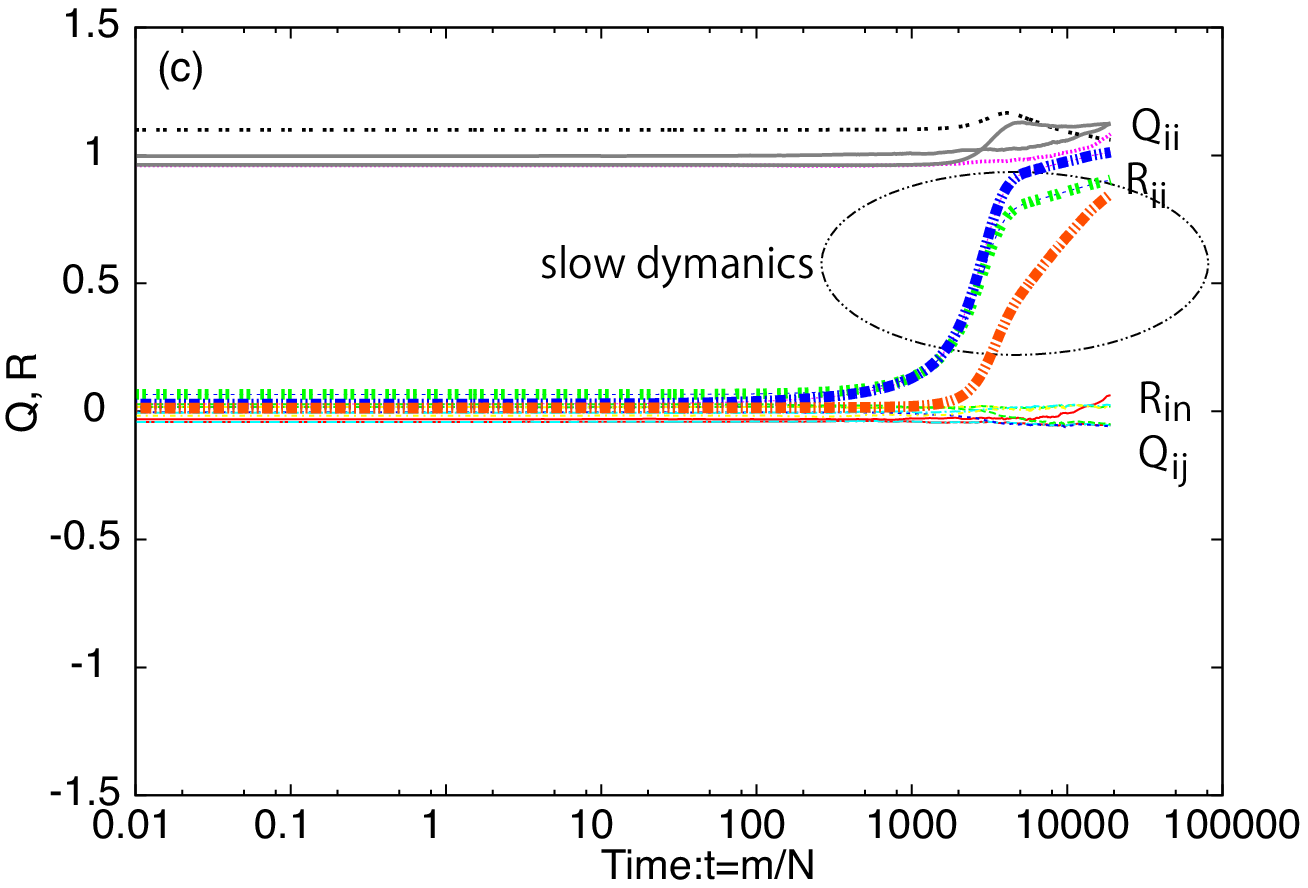}
\end{center}
\caption{\label{mse4}Dynamic behavior of SGD when teacher is singular. Network structures are $M=2$ and $K=4$. (a) shows MSE, (b) shows $\bm{w}$, and (c) shows $Q$s and $R$s.}
\end{figure}

Figure \ref{mse4}(a) shows the time course of the MSE, Fig. \ref{mse4}(b) shows that of the hidden-output weight $\bm{w}$, and Fig. \ref{mse4}(c) shows that of $Q$s and $R$s for SGD. The simulation conditions are the same as those in Fig. \ref{mse}. The teacher is $M=2$, and the student is $K=4$. From Fig. \ref{mse4}(a), the baseline of the MSE is a little bit larger than that in Fig. \ref{mse}(a) using the normal teacher. In Fig. \ref{mse4}(c), we can observe the slow dynamics of $R_{ii}$ shown by the broken circle. These slow dynamics are significant when $R_{ii}$ approaches $R_{ii}=1$, which is the singular point. 

\begin{figure}[ht]
\begin{center}
\includegraphics[width=6cm,height=2.5cm]{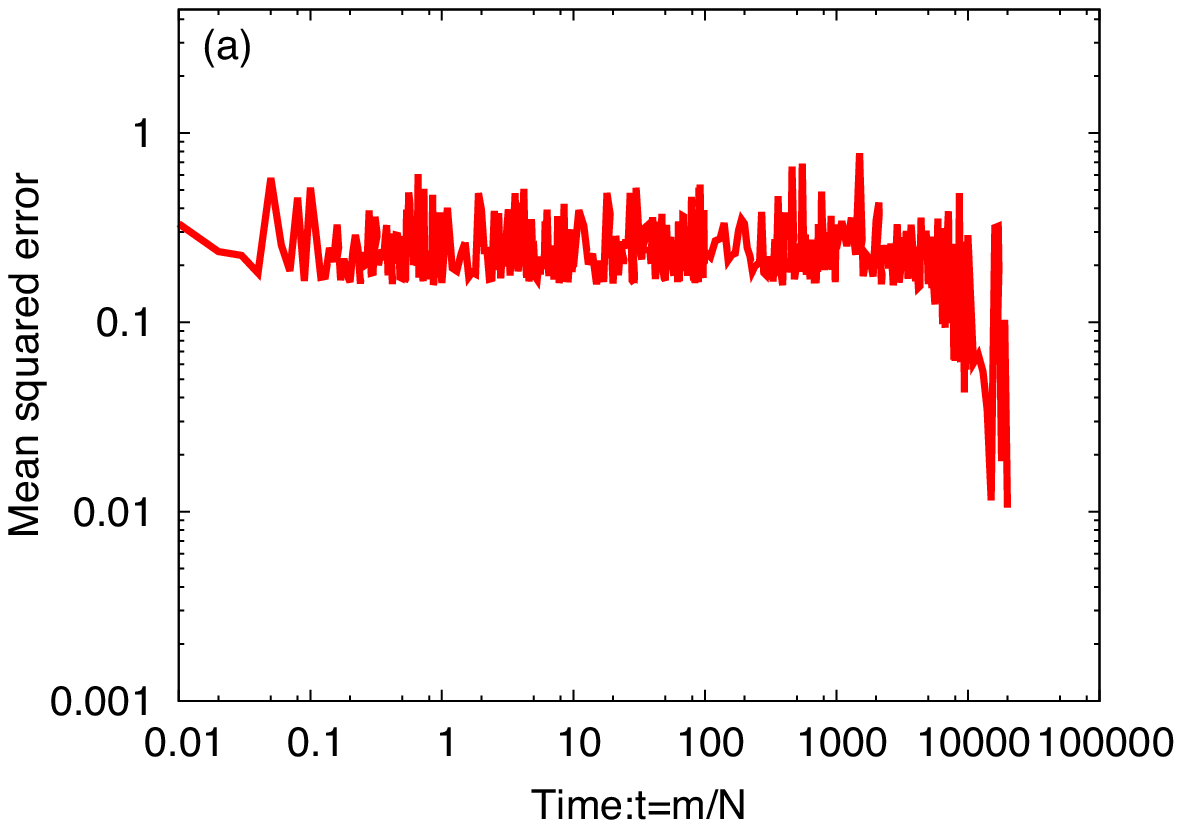}\\
\hspace{0.3cm}\includegraphics[width=5.6cm,height=2.4cm]{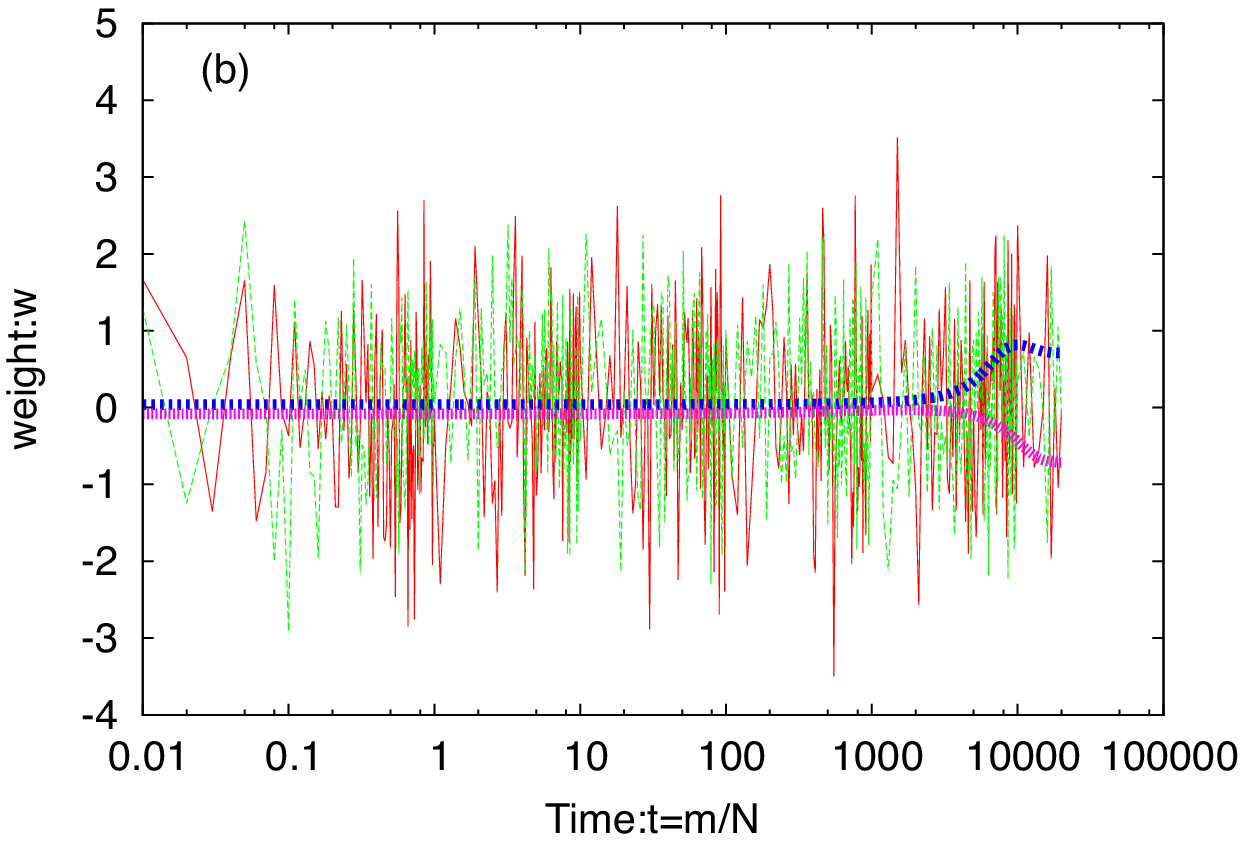}\\\includegraphics[width=6cm,height=4cm]{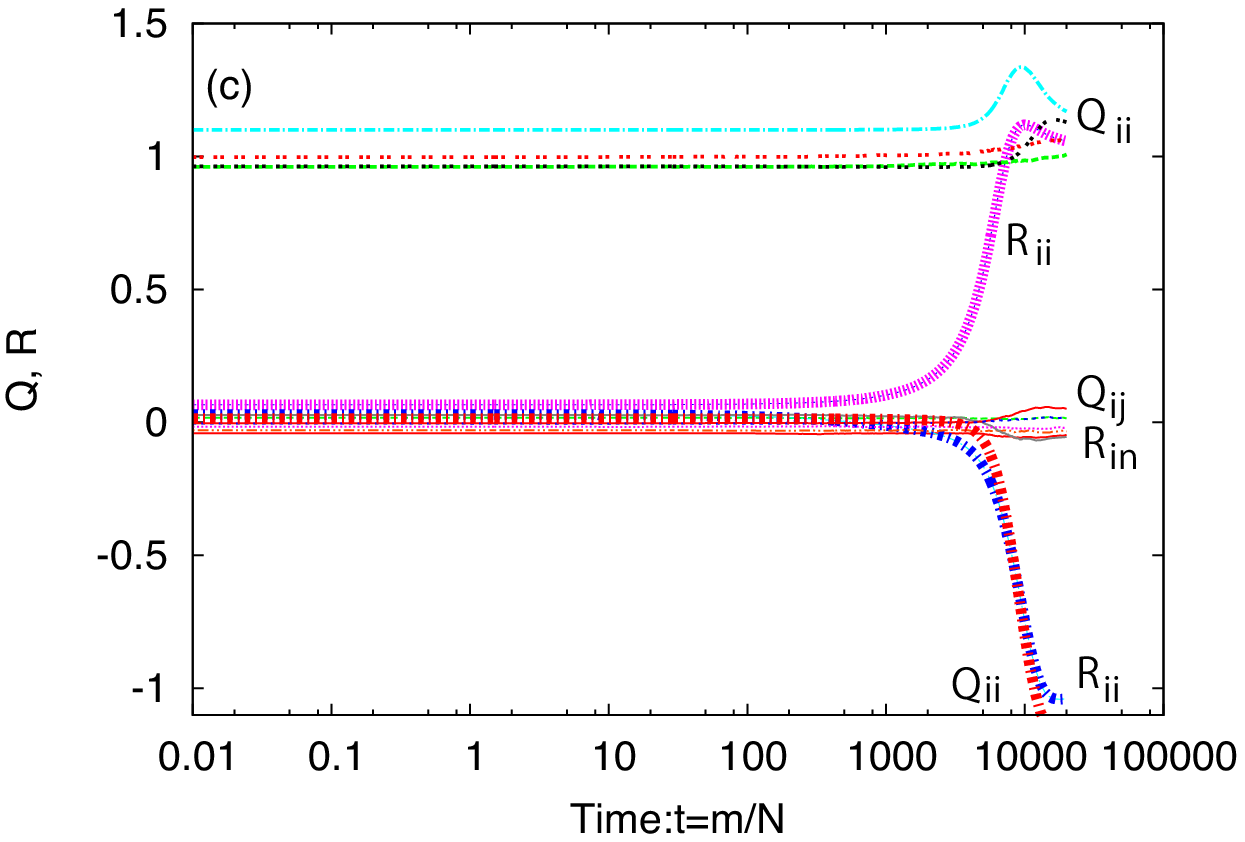}
\end{center}
\caption{\label{mse5}Dynamic behavior of dropout when teacher is singular. Network structures are $M=2$ and $K=4$. (a) shows MSE, (b) shows $\bm{w}$, and (c) shows $Q$s and $R$s.}
\end{figure}

Figure \ref{mse5}(a) shows the time course of the MSE, Fig.\ref{mse5}(b) shows that of the hidden-to-output weight $\bm{w}$, and Fig. \ref{mse5}(c) shows that of $Q$s and $R$s for dropout. The simulation conditions are the same as those in Fig. \ref{mse}. The teacher is $M=2$, and the student is $K=4$. From Fig. \ref{mse5}(a),  the baseline of MSE is small compared to that of Fig. \ref{mse4}(a). We can also observe the symmetry break of weights near $t=10,000$. From Fig. \ref{mse5}(c), $R_{ii}$ converged into $R_{ii}=1$ or $R_{ii}=-1$ rapidly near $t=10,000$, and we cannot observe slow dynamics around the singular point, that is $R_{ii}=1$. 

These results enable us to claim that by using dropout, the slow dynamics of $R$s may not be affected by the singular point. 

\section{Conclusion}
This paper presented our analysis of the behavior of dropout in online learning. In online learning, overfitting does not occur. Thus, we analyzed the dropout from other aspects: the learning behavior of students having redundant units and the learning behavior of students when the teacher is a singular. For redundant students, SGD cannot eliminate the redundant weights, and this causes large MSE baselines. However, dropout can eliminate the redundant weights properly and can achieve a small MSE baseline and also the occurrence of symmetry breaking of weights. For a singular teacher, SGD shows the slow dynamics of $R_{ii}$ near the singular point, that is, $R_{ii}=1$. However, dropout has not shown slow dynamics near the singular point, and it converges into a singular point rapidly. We have analyzed dropout in online learning through computer simulations. Our next step is the analysis of dropout using theoretical methods. 

\end{document}